\documentclass{article}

\usepackage{microtype}
\usepackage{graphicx}
\usepackage{subcaption}
\usepackage{multirow}
\usepackage{booktabs} % for professional tables
\usepackage{paralist}
\usepackage{adjustbox}
\usepackage[table]{xcolor}
\usepackage{paralist}

\usepackage{hyperref}

\usepackage[preprint]{icml2026}

\usepackage{amsmath}
\usepackage{amssymb}
\usepackage{mathtools}
\usepackage{amsthm}

\usepackage[capitalize,noabbrev]{cleveref}

\theoremstyle{plain}

\theoremstyle{definition}

\theoremstyle{remark}

\definecolor{mygray}{rgb}{0.9, 0.95, 1}

\usepackage[textsize=tiny]{todonotes}

\icmltitlerunning{MPF-Net}

\begin{document}

\twocolumn[
  \icmltitle{MPF-Net: Exposing High-Fidelity AI-Generated Video Forgeries via \\ Hierarchical Manifold Deviation and Micro-Temporal Fluctuations}



  \icmlsetsymbol{equal}{*}

  \begin{icmlauthorlist}
    \icmlauthor{Xinan He}{equal,ncu,szu}
    \icmlauthor{Kaiqing Lin}{equal,szu}
    \icmlauthor{Yue Zhou}{equal,szu}
    \icmlauthor{Jiaming Zhong}{szu}
    \icmlauthor{Wei Ye}{hst}
    \icmlauthor{Wenhui Yi}{ncu}
    \icmlauthor{Bing Fan}{nt}
    \icmlauthor{Feng Ding}{ncu}
    \icmlauthor{Haodong Li}{szu}
    \icmlauthor{Bo Cao}{cb}
    \icmlauthor{Bin Li}{szu}
  \end{icmlauthorlist}

  \icmlaffiliation{ncu}{Nanchang University}
  \icmlaffiliation{szu}{Guangdong Provincial Key Laboratory of Intelligent Information Processing, Shenzhen Key Laboratory of Media Security, and SZU AFS Joint Innovation Center for AI Technology, Shenzhen University}
  \icmlaffiliation{hst}{Huazhong University of Science and Technology, WuHan}
    \icmlaffiliation{nt}{University of North Texas}
% \icmlaffiliation{ten}{Tencent Youtu Lab}
\icmlaffiliation{cb}{The Smart City Research institute of China Electronics Technology Group Corporation, shenzhen}
    
  \icmlcorrespondingauthor{Bin Li}{libin@szu.edu.cn}
  \icmlcorrespondingauthor{Feng Ding}{fengding@ncu.edu.cn}

  \icmlkeywords{AIGC Forensics, AI-generated Video Detection, low-level vision, Deep Learning}

  \vskip 0.3in
]



\printAffiliationsAndNotice{}  % no special notice (required even if empty)

\begin{abstract}
  With the rapid advancement of video generation models such as Veo and Wan, the visual quality of synthetic content has reached a level where macro-level semantic errors and temporal inconsistencies are no longer prominent. However, this does not imply that the distinction between real and cutting-edge high-fidelity fake is untraceable. We argue that AI-generated videos are essentially products of a manifold-fitting process rather than a physical recording. Consequently, the pixel composition logic of consecutive adjacent frames residual in AI videos exhibits a structured and homogenous characteristic. We term this phenomenon \textbf{`Manifold Projection Fluctuations' (MPF)}. Driven by this insight, we propose a hierarchical dual-path framework that operates as a sequential filtering process. The first, \textbf{the Static Manifold Deviation Branch}, leverages the refined perceptual boundaries of Large-Scale Vision Foundation Models (VFMs) to capture residual spatial anomalies or physical violations that deviate from the natural real-world manifold (off-manifold). For the remaining high-fidelity videos that successfully reside on-manifold and evade spatial detection, we introduce \textbf{the Micro-Temporal Fluctuation Branch} as a secondary, fine-grained filter. By analyzing the structured MPF that persists even in visually perfect sequences, our framework ensures that forgeries are exposed regardless of whether they manifest as global real-world manifold deviations or subtle computational fingerprints.
  
\end{abstract}

\vspace{-1mm}
\section{Introduction}
\vspace{-1mm}

\begin{figure}[tb]
    \centering
  \includegraphics[width = 0.48\textwidth]{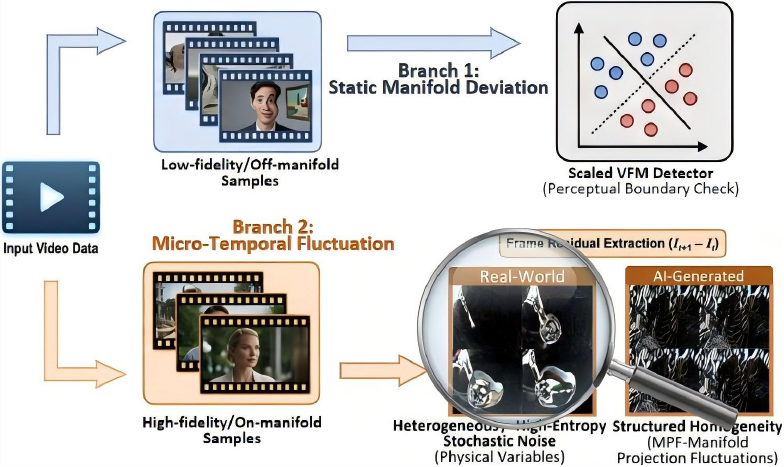}
  \caption{Proposed hierarchical dual-path forensic framework based on manifold and MPF Analysis.}
  \label{fig:outline_manifold}
  \vspace{-15pt}
\end{figure}

The recent surge in high-fidelity video generation models, exemplified by Sora2, Veo, and Wan, has pushed the quality boundaries of synthetic media to an unprecedented level. By achieving long-range temporal coherence and cinematic realism, synthetic content is now reaching a stage where macro-level semantic errors are no longer prominent, posing an immense challenge to digital forensics as fake videos become virtually indistinguishable from reality through content logic alone.

Earlier forensic research \cite{interno2025ai,zhang2025physics} primarily focused on identifying blatant physical law violations and semantic distortions resulting from the sparse fitting of generative models on the data manifold. However, we argue that the key to modern forensics lies not in these transient defects, which are easily `smoothed out' through model scaling, but in the unavoidable intrinsic artifacts embedded within the generation process itself.

The fundamental distinction between real-world content and AI-generated video: real-world content is an authentic recording of physical laws, while AI video is a computational mapping of a learned manifold. We project this discrepancy onto the frame residual features of consecutive frames to expose the forgery. In real videos, frame residuals maintain high pixel stability due to the constant photon counts on the sensor, yet remain subject to uncontrollable physical variables such as mechanical jitter and light refraction. Consequently, real-world frame residuals are \textbf{unstructured and heterogeneous}, characterized by high-entropy stochastic noise. In contrast, AI video generation relies on temporal conditioning and historical frame knowledge. Regardless of whether the content is on or off the natural manifold, the reconstruction logic is governed by fixed model weights, leading to \textbf{structured and homogenous} residual patterns. We define this phenomenon as Manifold Projection Fluctuations (MPF).

However, the efficacy of probing these micro-temporal artifacts is highly correlated with the video’s frame rate (FPS) and generation quality. For `off-manifold' videos, those characterized by low FPS, poor resolution, and significant high-level semantic distortions, the MPF signals are often obscured by catastrophic perturbations.

Inspired by the findings of \cite{zhou2025brought, veeramachaneni2025leveraging}, we acknowledge that Large-Scale Vision Foundation Models (VFMs) possess an inherent sensitivity to the distribution of the real-world manifold. As the training data of a VFM becomes more extensive, its perceptual boundaries for identifying synthetic patterns become increasingly refined. These models can efficiently `deviate' off-manifold samples by leveraging their dense internal projection of the natural world, even when temporal signals are corrupted. 

\begin{figure}[tb]
    \centering
  \includegraphics[width = 0.45\textwidth]{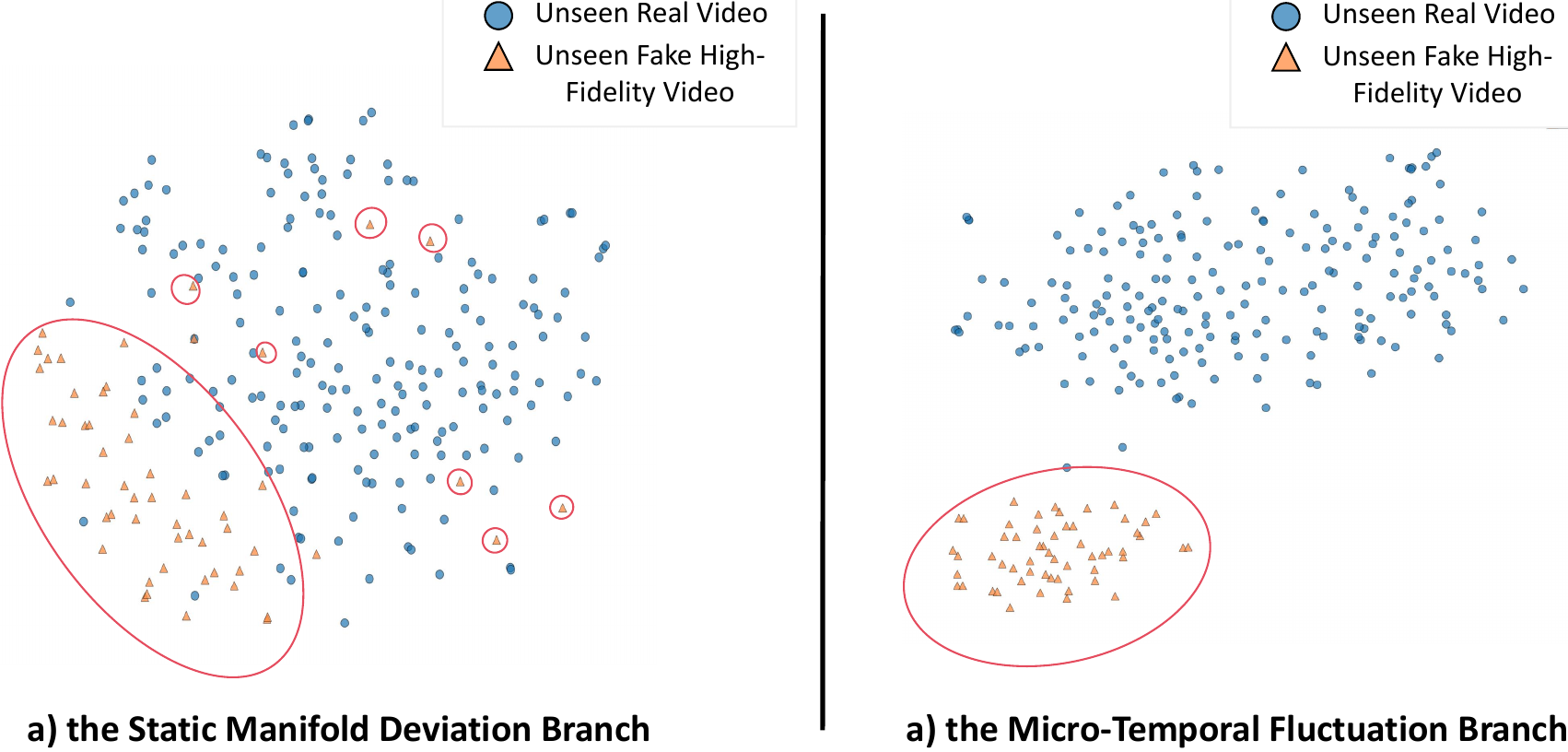}
  \caption{T-SNE visualization of the hierarchical feature spaces. In Branch I (left), while off-manifold samples with significant distortions are clearly deviated, high-fidelity `on-manifold' samples heavily overlap with the real-world distribution. In Branch II (right), the same on-manifold samples exhibit distinct, separable clustering when projected into the MPF-based feature space.}
  \label{fig:tsne}
  \vspace{-15pt}
\end{figure}

Building upon these insights, we propose a hierarchical dual-path framework that operates as a sequential filtering process. We categorize the current landscape of AI videos into two distinct branches:
\begin{compactenum}
    \item  \textbf{Branch I: Static Manifold Deviation}. For low-quality, off-manifold samples exhibiting clear semantic or spatial anomalies, we deploy a branch that leverages the refined perceptual boundaries of scaled VFMs. This stage acts as a `sentinel', efficiently capturing forgeries that deviate from the natural data distribution.
    \item \textbf{Branch II: Micro-Temporal Fluctuation }. For high-fidelity, on-manifold samples that flawlessly emulate spatial realism, we introduce a second branch dedicated to MPF analysis. By scrutinizing the micro-temporal residuals for structured homogeneity, this `microscope' stage exposes the deterministic computational DNA that spatial-only detectors fail to perceive.
\end{compactenum}

We validate this hierarchical logic through t-SNE visualizations of the feature spaces (ref to Fig.~\ref{fig:tsne}). In Branch I, while the VFM effectively isolates videos with obvious semantic distortions (off-manifold), there remains a significant overlap between high-fidelity synthetic samples and authentic data. In Branch II, however, the same high-fidelity samples when projected through our MPF-based feature space exhibit clear, separable clustering.
% Furthermore, to address the scarcity of high-fidelity and high-FPS samples in existing benchmarks, we introduce the High-Fidelity, In-the-Wild (HF-ITW) dataset (ref to Sec.~\ref{sec:HF-ITW Dataset Construction}). Comprising cinematic content from state-of-the-art APIs and hyper-realistic digital humans, this dataset focuses on visually indistinguishable `on-manifold' forgeries that lack detectable semantic flaws.

\vspace{-1mm}
\section{Related Work}
\vspace{-1mm}

\begin{figure*}[tb]
    \centering
  \includegraphics[width = 0.95\textwidth]{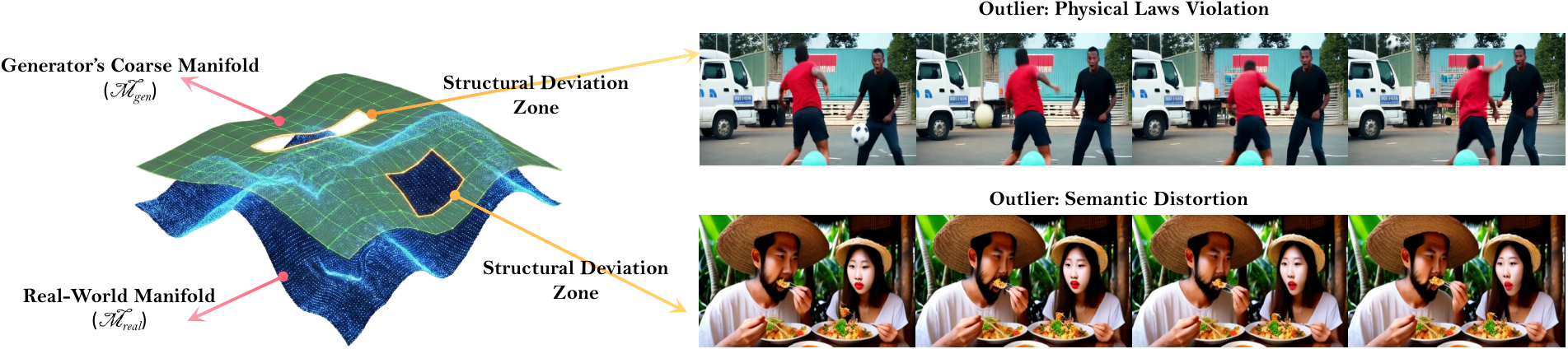}
  \caption{Synthetic Videos that lies on Generator's Coarse Manifold ($M_{Gen}$) but deviate from the structure of Real-World Manifold ($M_{real}$).}
  \label{fig:outline_manifold}
  \vspace{-15pt}
\end{figure*}

Video forgery detection has evolved through several distinct paradigms, shifting from high-level visual inspection to the analysis of underlying physics. \\ \textbf{Detection via Semantic Distortions and Inconsistencies.} Early forensic methodologies primarily focused on identifying spatial semantic distortions and temporal inconsistencies. DeCof \cite{ma2025detecting} demonstrated that capturing temporal artifacts offers superior generalizability compared to purely spatial analysis. HumanSAM \cite{liu2025humansam} further refined this by categorizing anomalies into spatial, appearance, and motion dimensions. To highlight spatial and temporal anomalies, AIGVDet \cite{bai2024ai} introduced optical flow analysis to expose motion incoherence, while Corvi et al. \cite{corvi2025seeing} utilized the 3D Fourier transform of video residuals to identify distinctive spectral peaks in synthetic content. Recent approaches have integrated Vision-Language Models (VLMs) to emulate human reasoning. For instance, Skyra \cite{li2025skyra} actively searches for intrinsic, physics-violating artifacts. However, with architectural innovations like Diffusion Transformers (DiTs) and the explosion of high-quality training data, models such as Wan and Veo have largely eliminated these macro-level semantic distortions, rendering content-based reasoning increasingly obsolete. \\
\textbf{Detection via Physical Constraints and Curvature.} To move beyond fragile semantic features, researchers have begun investigating the intrinsic physical constraints that distinguish real-world captures from generative fits. NSG-VD \cite{zhang2025physics} proposed the Normalized Spatiotemporal Gradient (NSG) based on the principle of probability flow conservation, utilizing pre-trained diffusion models to estimate discrepancies between spatial gradients and temporal density changes. Based on the `perceptual straightening' hypothesis, ResTraV \cite{interno2025ai} calculates the geometric curvature and distance of temporal trajectories within the frozen DINOv2 feature space. The core philosophy of these methods is that AI fits statistical distributions rather than simulating the deterministic physical paths of light. Nevertheless, as generative models are trained on increasingly vast datasets, they have begun to statistically perfectly imitate these physical laws. Much like semantic distortions, these physical anomalies are gradually being `smoothed out' by iterative model improvements. 

% Unlike existing works that rely on semantic reasoning or empirical physical laws, our approach targets the Manifold Projection Fluctuations (MPF) inherent in the pixel-composition logic of generative models. By combining a manifold-aware frame detector with micro-temporal residual analysis, we expose the structured, homogenous fingerprints of the AI generation process, achieving state-of-the-art performance in high-fidelity video forensics.

\vspace{-1mm}
\section{The `Gun' in Spatial Forensics: VFM as a Manifold Sentinel}
\vspace{-1mm}

Earily video generative models suffered from `sparse fitting' due to limited parameter scales and training data. When confronted with complex logic, rare object combinations, or counter-intuitive physical prompts, these models encounter regions of the manifold where they lack sufficient structural reference. In these data-sparse zones, the model is forced into erroneous inference, projecting samples into regions far from the natural distribution. These projections manifest as what we traditionally term semantic distortions or physical laws violations. In the manifold perspective, these are off-manifold outliers that fail to align with the underlying geometry of reality, as show in Fig.~\ref{fig:outline_manifold}.

As argued by Zhou et al. \cite{zhou2025brought}, advanced Vision Foundation Models (VFMs) act as a `gun' in this forensic `knife fight'. Unlike previous expert detectors trained on specific artifacts, a state-of-the-art VFM (e.g., MetaCLIP2) has internalized a vastly more comprehensive and dense representation of $\mathcal{M}_{real}$ through internet-scale pre-training.

Therefore, leveraging Large-Scale VFMs as our Manifold Sentinel enables the effective identification of off-manifold samples characterized by semantic distortions and physical law violations. Their architectures possess rigorous perceptual boundaries that capture subtle semantic inconsistencies and statistical anomalies, to to discern structural discrepancies within $\mathcal{M}_{real}$ that remain imperceptible to narrower, task-specific models. 

\vspace{-1mm}
\section{The Residual `Blind Spot': the Gun is Not Enough}
\vspace{-1mm}
\label{sec:gunisnotenough}

\begin{figure*}[tb]
    \centering
  \includegraphics[width = 0.95\textwidth]{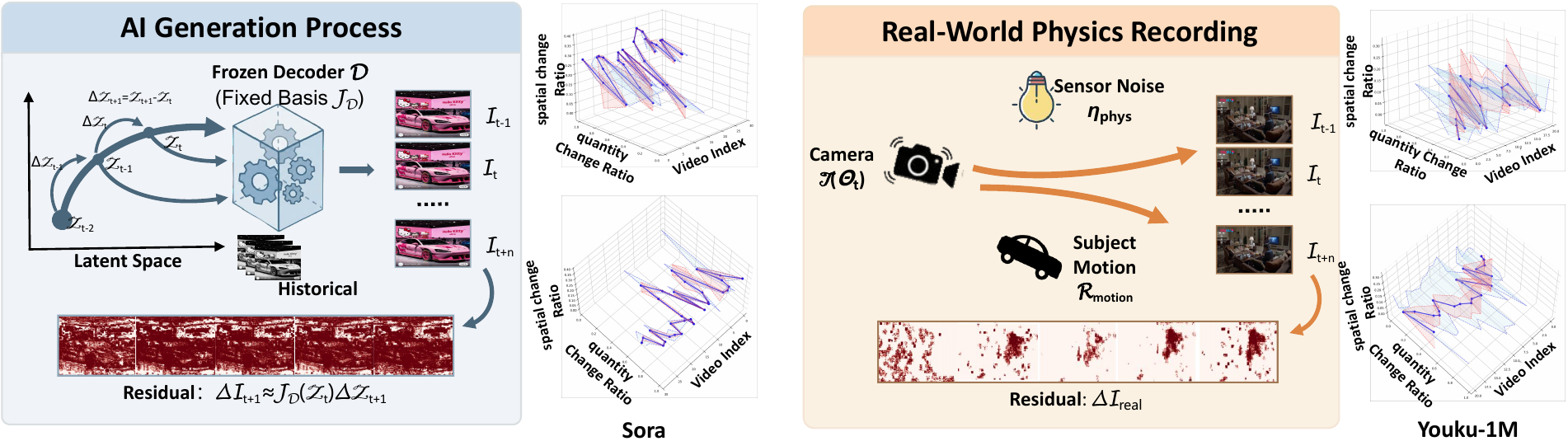}
  \caption{Micro-Temporal Fluctuation (MPF) Analysis, contrasts the homogenous, structured residuals of AI synthesis, constrained by a frozen decoder's fixed basis (left), with the heterogeneous, unstructured noise inherent in physical world recordings (right).}
  \label{fig:outline_mpf}
  \vspace{-15pt}
\end{figure*}

Despite advanced VFMs' spatial prowess, they exhibit a significant blind spot when tasked with identifying high-fidelity videos that flawlessly fit the intrinsic structure of the real-world manifold. To bridge this forensic gap, we introduce the analysis of Micro-Temporal Fluctuation (MPF) phenomena.

\textbf{\underline{The Mimicry of Temporal Conditioning}}: As shown in Fig.~\ref{fig:outline_mpf} left, Unlike AI image generation, which often produces isolated samples, modern AI video models leverage deep temporal conditioning. Each frame is generated with guidance from historical frames ($I_{t-1}, I_{t-2}, \dots$), which act as a strong semantic anchor. This temporal `pull' ensures that generated pixels are not only spatially plausible but are forced to adhere to the high-level continuity of the natural data manifold.

We formalize the generation process within a short temporal window as a manifold projection. Let $I \in \mathbb{R}^N$ denote a single frame in the pixel space where $N = H \times W \times C$. This frame is projected from a dense latent representation $z \in \mathbb{R}^M$ via a pre-trained, frozen decoder $\mathcal{D}$:

\vspace{-2mm}
\begin{equation}
I_t = \mathcal{D}(z_t).
\end{equation}

The latent vector $z \in \mathbb{R}^M$ serves as the intrinsic nexus of the generation, capturing the necessary information for the current window: historical hidden states ($I_{<t}$), text prompt embeddings, and the deterministic trajectories formed after the diffusion denoising process.

In high-fidelity videos, the transition between $t$ and $t+1$ is driven by a smooth shift $\Delta z = z_{t+1} - z_t$ within the latent manifold. Applying a first-order Taylor expansion, the resulting pixel-level residual $\Delta I$ (the MPF) can be approximated as:

\vspace{-2mm}
\begin{equation}
    \Delta I_{t+1} = \mathcal{D}(z_t + \Delta z_{t+1}) - \mathcal{D}(z_t) \approx \mathbf{J}_\mathcal{D}(z_t) \cdot \Delta z_{t+1},
\end{equation}

where $\mathbf{J}_\mathcal{D}(z) \in \mathbb{R}^{N \times M}$ is the Jacobian Matrix of the decoder:

\vspace{-2mm}
\begin{equation}
\mathbf{J}_\mathcal{D}(z) = \frac{\partial \mathcal{D}}{\partial z} = \begin{bmatrix} \frac{\partial I_1}{\partial z_1} & \cdots & \frac{\partial I_1}{\partial z_M} \\ \vdots & \ddots & \vdots \\ \frac{\partial I_N}{\partial z_1} & \cdots & \frac{\partial I_N}{\partial z_M} \end{bmatrix},
\end{equation}
where each row represents the sensitivity of a specific pixel to the entire set of latent features, each column defines a spatial perturbation map. It dictates how all pixels in the frame must move in coordination when a single latent feature $z_j$ changes.

This formulation reveals the fundamental discrepancy between AI synthesis and physical recording through two key properties:
\begin{compactenum}
 \item \textbf{Fixed Basis Functions}: The parameters of the decoder $\mathcal{D}$ are frozen during inference. Consequently, the Jacobian matrix $\mathbf{J}_\mathcal{D}$ acts as a static set of basis functions that define the permissible directions of pixel fluctuations. Every temporal transition $\Delta I$ is merely a linear combination of these $M$ basis vectors.
 \item \textbf{Structural Homogeneity}: Within a short temporal window, $z$ moves slowly along the manifold due to high semantic overlap between frames. This implies that $\mathbf{J}_\mathcal{D}(z)$ exhibits high local smoothness ($\mathbf{J}_\mathcal{D}(z_t) \approx \mathbf{J}_\mathcal{D}(z_{t+1})$). Consequently, the model repeatedly utilizes the same spatial perturbation patterns to reconstruct the sequence. This results in temporal residuals that are highly structured and predictable.
\end{compactenum}

\textbf{\underline{The Physics of Reality}}: As shown in Fig.~\ref{fig:outline_mpf} right, for the real-world video, the difference between consecutive frames can be formalized as:
\vspace{-1mm}
\begin{equation}
    \Delta I_{real} = \mathcal{T}(\theta_t, I_t) + \eta_{phys} + \mathcal{R}_{motion},
\end{equation}
where $\mathcal{T}(\theta_t, \cdot)$ represents transformations driven by physical parameters $\theta_t$ (e.g., camera jitter, rotation), stochastic extrinsic factors $\eta_{phys}$ (e.g., sensor shot noise, ambient lighting fluctuations), and $\mathcal{R}_{motion}$ is the residual caused by subject motion within the scene.

\textbf{\underline{Conclusion}}: Therefore, This physics-based generation exhibits two fundamental forensic properties:

\begin{compactenum}
    \item \textbf{Spatial Sparsity}. In high-fidelity stable regions where $\mathcal{T} \to 0$ and $\eta_{phys} \to 0$ (e.g., stationary backgrounds), real-world pixels achieve absolute stability ($\Delta I_{bg} \approx 0$). Unlike AI-generated MPF, which manifests as persistent, structured fluctuations due to continuous latent drift, authentic residuals are spatially sparse and vanish in the absence of physical change.
    \item \textbf{Temporal Heterogeneity}. Authentic noise $\eta_{phys}$ is inherently high-entropy and temporally decorrelated. Any introduced physical variable ($\mathcal{T}$ or $\eta_{phys}$) produces residuals that are unstructured and heterogeneous across both space and time. Because these signals lack a shared, frozen basis (the Jacobian $\mathbf{J}_\mathcal{D}$), they do not exhibit the structural homogeneity or repetitive perturbation patterns characteristic of AI manifold projections.
\end{compactenum}

Futhermore, we visualize the discrepancy between AI-generated (Sora) and authentic (Youku-mPLUG) videos within a macroscopic 3D Feature Space (refer to Fig.~\ref{fig:outline_mpf}). This visualization maps each video's spatio-temporal signal patterns across three critical dimensions: \emph{X-axis (Temporal Index)}: Represents the sequential video samples. \emph{Y-axis (Quantity Mean)}: The mean magnitude of pixel-wise changes. \emph{Z-axis (Spatial Mean)}: The mean spatial distribution of these changes. As shown in the figure, high-fidelity AI videos manifest as a tight, linear trajectory with minimal variance. The fluctuations in both quantity (indicated by the \textcolor{red}{red} deviation lines) and spatial distribution (\textcolor{blue}{blue} deviation lines) are highly constrained. Conversely, real-world videos exhibit Volatility, forming a dispersed and chaotic cloud in the feature space.
 
\vspace{-1mm}
\section{Hierarchical Dual-path Framework}
\vspace{-1mm}

\begin{figure*}[tb]
    \centering
  \includegraphics[width = 0.93\textwidth]{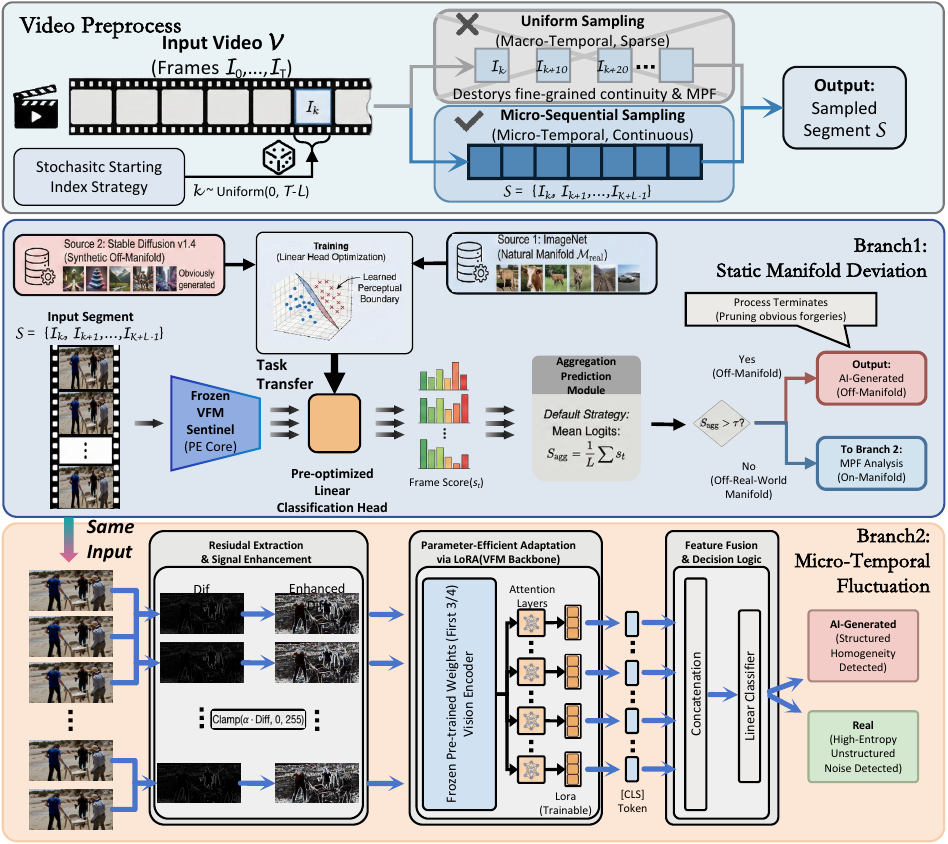}
  \caption{The hierarchical dual-path
framework: Video Preprocessing, Static Manifold Deviation Branch and Micro-Temporal Fluctuation.}
  \label{fig:outline_branch}
  \vspace{-10pt}
\end{figure*}

The framework is structured to address two distinct categories of forgeries: off-manifold and on-manifold samples. The overall detection logic is formalized as a sequential decision process: 1) the Static Manifold Deviation Branch: utilizes a VFM-based Manifold Sentinel to perform a distribution check on individual frames. 2) the Micro-Temporal Fluctuation Branch: focuses on the Manifold Projection Fluctuations (MPF) hidden within the frame residuals.

\textbf{Video Preprocessing (ref to Fig.~\ref{fig:outline_branch} top).} Traditional forensics often aim to capture high-level motion anomalies or long-range semantic inconsistencies. Therefore, they almost focus on macro-temporal patterns through uniform frame sampling. We argue that uniform sampling, which extracts frames at fixed intervals (e.g., every 10th frame) is counterproductive for identifying high-fidelity forgeries, as it destroys the fine-grained continuity required to observe the MPF.

To preserve these subtle traces, we abandon uniform sampling in favor of Micro-Sequential Sampling. Given a video $V$ consisting of $T$ frames, instead of sparse extraction, we isolate a continuous segment of $L$ frames:
\vspace{-1mm}
\begin{equation}
    S = \{I_k, I_{k+1}, \dots, I_{k+L-1}\}.
\end{equation}
This ensures that the Jacobian matrix $\mathbf{J}_{\mathcal{D}}$ remains locally smooth across the segment, allowing the structured homogeneity of the reconstruction error to manifest clearly in the residuals $\Delta I_t = I_{t+1} - I_t$.

To ensure the robustness of our detector and avoid over-fitting to specific temporal phases (such as video intros or outros), we implement a Stochastic Starting Index strategy. For each training and inference pass, the starting frame $k$ is randomly sampled:
\vspace{-1mm}
\begin{equation}
    k \sim \text{Uniform}(0, T - L),
\end{equation}
where $T$ is the total number of frames and $L$ is the segment length. However, to ensure experimental reproducibility and standardized evaluation across different baselines, we adopt a fixed indexing strategy during the inference (testing) phase. Specifically, the segment extraction consistently begins from the initial frame ($k=0$).

\subsection{The Static Manifold Deviation Branch}
\vspace{-1mm}
This branch serves as the initial Sentinel stage of our framework. Its primary objective is to evaluate individual video frames against the real-world data manifold and isolate off-manifold synthetic content that exhibits spatial anomalies, semantic distortions, or physical laws violations  (ref to Fig.~\ref{fig:outline_branch} middle).

\textbf{Latent Representation via State-of-the-Art VFM}. To establish a robust manifold sentinel, we adopt the pretrained latent space of the PE Core \cite{bolya2025perception} as our primary representation framework, prioritizing its unique capacity for capturing high-fidelity visual structures. Furthermore, the model’s extensive pre-training on high-quality, large-scale video-text datasets provides it with an exceptionally dense and nuanced understanding of the real-world data manifold ($\mathcal{M}_{real}$).

\textbf{Establishing the Boundary for Off-Manifold Deviation}. To identify a robust boundary line that separates off-manifold synthetic content from natural images, we leverage a pre-optimized linear classification head transferred from established AIGI (AI-Generated Image) forensic tasks. Rather than training a new classifier from scratch on specific video data, we incorporate learned priors derived from a foundational dataset pair \cite{zhu2023genimage}: ImageNet (representing the natural real-world manifold $\mathcal{M}_{real}$) and Stable Diffusion (SD) v1.4 (representing synthetic off-manifold samples). This combination has been widely proven in existing works to yield the best generalization performance for high-level AIGC detection. By fine-tuning only the linear head, we establish a stable `perceptual boundary' that captures the high-level semantic manifold structure of AI-generated content without over-fitting to specific video generators.

\textbf{Frame-level Inference and Sequential Aggregation}. During inference, the continuous segment of $S$ frames extracted is processed through the frozen PE Core. Each frame $I_t$ is mapped to the latent space to generate a feature vector, which is then fed into the pre-trained linear head to obtain a logit score $s_t$. To determine the overall status of the video segment, we introduce an Aggregation Prediction Module. We evaluated three distinct aggregation strategies: \underline{Mean Logits}. Calculating the average score across the sequence: $S_{agg} = \frac{1}{L} \sum_{t=1}^{L} s_t$. 

% \underline{1) Majority Voting}:The final classification of the video segment is determined by the majority of individual frame predictions. If more than $L/2$ frames are classified as off-manifold, the entire sequence is flagged as synthetic. \underline{2) Max Logits}:  The video's status is determined solely by the frame exhibiting the highest confidence or anomaly score ($s_{max} = \max\{s_1, \dots, s_L\}$). If even a single frame demonstrates a sufficiently high deviation from the manifold, the entire video is deemed a forgery. \underline{3) Mean Logits (Default)}: Calculating the average score across the sequence: $S_{agg} = \frac{1}{L} \sum_{t=1}^{L} s_t$. Experimental results indicate that all three strategies yield comparable performance. Consequently, we adopt the Mean Logits strategy as our default for its stability in smoothing out transient sensor noise.

The final output of the Static Branch serves as a gatekeeper for the entire framework: \underline{Case I}: Off-Manifold. If $S_{agg}$ exceeds a predefined threshold $\tau$, the video is classified as AI-generated. The process terminates here, achieving high efficiency by pruning obvious forgeries. \underline{Case II}: On-Real-World Manifold. If $S_{agg} \le \tau$, the video is considered to be on-manifold (visually indistinguishable from reality). These challenging samples are then passed to the Micro-Temporal Fluctuation Branch for a second-stage MPF analysis of their pixel-composition logic.

% \begin{figure*}[h!]
%     \centering
%   \includegraphics[width = 0.95\textwidth]{Figures/outline_branch2.pdf}
%   \caption{Micro-Temporal Fluctuation Branch Residual Extraction.}
%   \label{fig:outline_branch2}
%   \vspace{-10pt}
% \end{figure*}

\subsection{The Micro-Temporal Fluctuation Branch}
\label{sec:microtemporalbranch}
\vspace{-1mm}
This branch focuses exclusively on identifying MPF at the signal level  (ref to Fig.~\ref{fig:outline_branch} bottom).

\textbf{Residual Extraction and Signal Enhancement}. To expose the subtle MPF effect, we first compute the pixel-level differences between adjacent frames in the continuous sequence $S$. For each pair of adjacent frames $(I_t, I_{t+1})$, we calculate the absolute residual $\Delta I_t$ and apply a linear normalization and clamping procedure to enhance the faint generative signals:
\vspace{-1mm}
\begin{equation}
    \Delta I_t = \text{Clamp}(\alpha \cdot |I_{t+1} - I_t|, 0, 255),
\end{equation}
where $\alpha = 10.0$ is the scaling factor used to amplify micro-fluctuations. This normalization process highlights the pixel stability in identical regions across frames and exposes the structured homogeneity of the residual features. 

% \textbf{Backbone Selection: MetaCLIP as a Signal Microscope.} For processing the enhanced residual features, we employ MetaCLIP (specifically MetaCLIP-v2) as our backbone. While other Vision Foundation Models such as DINOv3, PE Core, or standard CLIP possess strong spatial reasoning, we select MetaCLIP for its architectural efficiency and superior fine-tuning sensitivity. Compared to larger VFMs, MetaCLIP is significantly more lightweight and exhibits a higher capacity for capturing low-level signal patterns when adapted for downstream forensic tasks

\textbf{Parameter-Efficient Adaptation via LoRA}. Since existing VFMs were primarily pre-trained for high-level semantic recognition rather than pixel-wise low-level semantic analysis, we must adapt the model to recognize MPF without destroying its intrinsic latent representation space. We implement LoRA, focusing exclusively on the attention layers of the final quarter (1/4) of the backbone. By freezing the majority of the pre-trained weights and only fine-tuning a small fraction of the parameters in the terminal layers, we enable the model to learn the specific `fingerprints' of MPF while preserving the robust feature-extraction capabilities of the VFM.

\textbf{Feature Fusion and Decision Logic}. To determine the final classification of the sequence, we extract the [CLS] token from the last layer of the VFM backbone for each enhanced residual frame. Our empirical analysis shows that simple concatenation of these [CLS] tokens provides superior generalization performance compared to introducing additional complex temporal modules (such as Temporal Transformers).

As the second stage of our hierarchical defense, this branch provides the definitive verdict for `on-manifold' samples. If the concatenated residual features exhibit the structured homogeneity of MPF, the video is flagged as AI-generated. If the residuals present the high-entropy, unstructured characteristics of physical recording, the video is classified as Real.

\begin{table}[t] % Use [h!] or similar for placement in appendix
  \centering
    \caption{Comparison of \emph{MPF-Net} vs. AIGC experts on VidProM \cite{wang2024vidprom}. $\uparrow$ Higher is better.}
    % \vspace{-5pt}
    \label{tab:vidprom} % Keep original label
  \footnotesize
  \setlength{\tabcolsep}{3pt}
  \begin{adjustbox}{width=\linewidth}
      \begin{tabular}{lccccc}
        \toprule
        & \multicolumn{5}{c}{\textbf{Accuracy ↑} (\%)}\\
        \cmidrule(lr){2-6}
        Method &
          Pika & VC2 & T2VZ & MS & Avg \\
        \midrule
        
        CNNSpot~\cite{wang2020cnn}
          & 0.5117 & 0.5018 & 0.4997 & 0.5031 & 0.5041 \\
        FreDect~\cite{doloriel2024frequency}
          & 0.5007 & 0.5403 & 0.6988 & 0.6994 & 0.6098\\
        Fusing~\cite{ju2022fusing}
          & 0.5060 & 0.5007 & 0.4981 & 0.5128 & 0.5044\\
        Gram-Net~\cite{liu2020global}
          & 0.8419 & 0.6742 & 0.5248 & 0.5046 & 0.6364\\
        GIA~\cite{frank2020leveraging}
          & 0.5373 & 0.5175 & 0.4105 & 0.6022 & 0.5169\\
        LNP~\cite{tan2023learning}
          & 0.4348 & 0.4510 & 0.4750 & 0.4521 & 0.4532\\
        DFD~\cite{bi2023detecting}
          & 0.5053 & 0.4995 & 0.4896 & 0.4832 & 0.4944\\
        UnivFD~\cite{ojha2023towards}
          & 0.4941 & 0.4865 & 0.4958 & 0.5743 & 0.5127\\
        ReStraV~\cite{interno2025ai}
          & 0.9090 & 0.9950 & 0.9905 & \textbf{0.9837} & \textbf{0.9706}\\
        \midrule
        \rowcolor{mygray}
        \textbf{MPF (Ours)} & \textbf{1.0000} & \textbf{0.9997} & \textbf{0.9938} & 0.8115 & 0.9513\\
        \bottomrule
      \end{tabular}
  \end{adjustbox}
    \vspace{-10pt}
\end{table}

\begin{table*}[t]
    \centering
    \caption{Comparison of \emph{MPF-Net} vs. AIGC experts on GenVideo~\cite{chen2024demamba}, with `PIKA' trainging condition. $\uparrow$ Higher is better.}
    \vspace{-5pt}
    \label{12many}
    \resizebox{0.98\textwidth}{!}{
    \begin{tabular}{ccccccccccccc}
    \toprule
    \multirow{3}{*}{Model}&\multirow{3}{*}{Metric}&\multicolumn{10}{c}{Testing Subset}&\multirow{3}{*}{Avg.}
    \\ \cline{3-13}
    
    &&\multirow{2}{*}{Sora}&Morph&\multirow{2}{*}{Gen2}&\multirow{2}{*}{HotShot}&\multirow{2}{*}{Lavie}&\multirow{2}{*}{Show-1}&Moon&\multirow{2}{*}{Crafter}&Model&Wild&\\
    &&&Studio&&&&&Valley&&Scope&Scrape\\
    \midrule

    \multirow{2}{*}{NPR~\cite{tan2024rethinking}}&
    R& 0.5536 & 0.7757& 0.7188&0.4860& 0.0721& 0.0429& 0.8626& 0.6029&0.7143& 0.3153& 0.5144\\
&F1&0.4627&0.8410&0.8197&0.0871&0.1293&0.0772&0.8894&0.7368&0.8019&0.4604&0.5306\\
    \midrule
    
    \multirow{2}{*}{STIL~\cite{gu2021spatiotemporal}}&
    R& 0.7500& 0.7943 & 0.9449 & 0.5786 & 0.5314 & 0.6414 & 0.9712 & 0.8529 & 0.6943 & 0.6242 &0.7383 \\
&F1&0.0984&0.5530&0.7580&0.4355&0.5128&0.4708&0.6107&0.7130&0.5008&0.5117&0.5165\\
    \midrule
    
    \multirow{2}{*}{MINTIME-CLIP-B~\cite{coccomini2024mintime}}
    &R&0.1607&0.2686&0.1471&0.0557&0.3757&0.0900&0.2188&0.4071&0.0114&0.4309 & 0.2166\\    &F1&0.2338&0.4178&0.2545&0.1039&0.5428&0.1626&0.3535&0.5752&0.0222&0.5969 & 0.3263\\
    \midrule
    
    \multirow{2}{*}{FTCN-CLIP-B~\cite{Zheng2021ExploringTC}}&
    R&0.9107&0.9300&0.9232&0.1771&0.1750&0.0643&0.9968&0.8479&0.7457&0.5173 & 0.6287\\    &F1&0.5896&0.9195&0.9371&0.2770&0.2855&0.1088&0.9483&0.8948&0.8106&0.6513 & 0.6423\\
    \midrule
    
    \multirow{2}{*}{TALL~\cite{xu2023tall}}&
    R& 0.6071& 0.7043 &0.9449  & 0.5914 &0.5757  &0.6400  &0.9633  & 0.8871 & 0.6057  &0.6210&0.7141 \\
    &F1&0.1038& 0.5609  &0.8029  &0.4932  &0.5817  &0.5231  & 0.6726  &  0.7746&0.5024  &0.5569& 0.5572\\
    \midrule 
    
    \multirow{2}{*}{XCLIP-B-FT~\cite{ni2022expanding}}&R& 0.6786 & 0.9129  & 0.9623 & 0.1200 & 0.2236 & 0.0914 & 0.9984 & 0.8343 & 0.7557 & 0.5184 &0.6096 \\
&F1&0.5891&0.9301&0.9679&0.2049&0.3581&0.1579&0.9712&0.8974&0.8361&0.6662&0.6579\\
    \midrule
    
    \multirow{2}{*}{DeMamba-XCLIP-FT~\cite{chen2024demamba}}&R &0.9286 & 0.9729 & 0.9848 & 0.3829 & 0.5350 & 0.4143 & 0.9984 & 0.9407 & 0.7729 & 0.6415 &0.7572 \\
&F1&0.4298&0.8990&0.9460&0.4864&0.6559&0.5160&0.9017&0.9235&0.7869&0.7183&
0.7263\\
    \midrule
    
    \multirow{2}{*}{ReStraV~\cite{interno2025ai}}&R&-&-&-&-&-&-&-&-&-&-&0.7350\\
    &F1&-&-&-&-&-&-&-&-&-&-&0.8270\\
    \midrule
    
    \multirow{2}{*}{NSG-VD~\cite{zhang2025physics}}&R&0.7857&0.9833&0.8000&\textbf{0.9250}&0.9417&\textbf{0.8750}&1.0000&0.9833&0.6833&0.8250&0.8802\\
&F1&0.8713&0.9833&0.8727&\textbf{0.9174}&0.9417&\textbf{0.9052}&0.9677&0.9593&0.7885&0.8800&0.9087\\
    \midrule

    \multirow{2}{*}{Skyra-RL-GenVideo(7B)~\cite{li2025skyra} }&R&\textbf{0.9564}&0.9586&0.9623&0.8271&\textbf{0.9429}&0.7200&0.9792&0.9807&0.6629&0.7863&0.8766\\
&F1&\textbf{0.9550}&0.9451&0.9530&0.8779&0.9432&0.8136&0.9616&0.9601&0.7676&0.8541&0.9000\\
    \midrule

    \rowcolor{mygray}&R&0.8393&\textbf{0.9900}&\textbf{0.9978}&0.8571&0.9079&0.8429&\textbf{1.0000}&\textbf{0.9993}&\textbf{0.9943}&\textbf{0.8348}&\textbf{0.9263}\\
    % real 0.9985
     \rowcolor{mygray}
     \multirow{-2}{*}{\textbf{MPF(Ours)}}&F1&0.7642&\textbf{0.9809}&\textbf{0.9917}&0.9091&\textbf{0.9446}&0.9008&\textbf{0.9843}&\textbf{0.9929}&\textbf{0.9831}&\textbf{0.8992}&\textbf{0.9351}\\
    
    \bottomrule
    \end{tabular}}
    \vspace{-10pt}
    \label{tab:genvideo_pika}
\end{table*}

\vspace{-1mm}
\section{Experiments}
\vspace{-1mm}

\subsection{Implementation Details}
\vspace{-1mm}
Our framework is implemented using PyTorch and executed on a workstation equipped with 8 NVIDIA L40 GPUs. However, we emphasize that both the training and inference phases can be comfortably completed on a single consumer-grade GPU.

\begin{figure*}[h!]
    \centering
  \includegraphics[width = 0.9\textwidth]{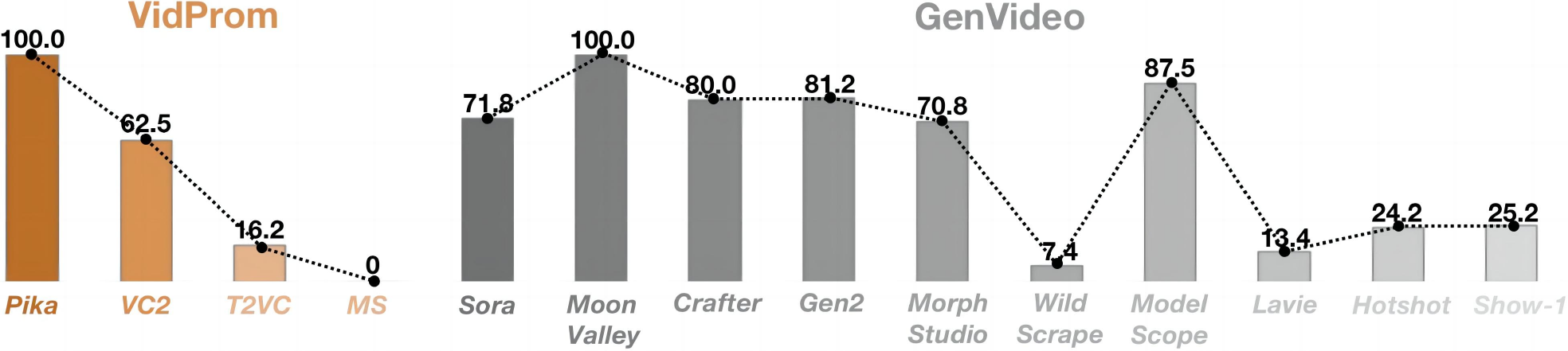}
  \caption{The efficacy of MPF across varying video quality spectra.}
  \label{fig:outline_MPFefficacy}
  \vspace{-10pt}
\end{figure*}

\begin{table*}[h!]
    \centering
    \caption{Comparison of \emph{MPF-Net} vs. AIGC experts on GenVideo \cite{chen2024demamba}, with `SENIE' trainging condition. $\uparrow$ Higher is better.}
    \vspace{-5pt}
    \label{12many}
    \resizebox{0.98\textwidth}{!}{
    \begin{tabular}{ccccccccccccc}
    \toprule
    \multirow{3}{*}{Model}&\multirow{3}{*}{Metric}&\multicolumn{10}{c}{Testing Subset}&\multirow{3}{*}{Avg.}
    \\ \cline{3-13}
    
    &&\multirow{2}{*}{Sora}&Morph&\multirow{2}{*}{Gen2}&\multirow{2}{*}{HotShot}&\multirow{2}{*}{Lavie}&\multirow{2}{*}{Show-1}&Moon&\multirow{2}{*}{Crafter}&Model&Wild&\\
    &&&Studio&&&&&Valley&&Scope&Scrape\\
    \midrule

    \multirow{2}{*}{NPR~\cite{tan2024rethinking}}&
    R&0.4643& 0.7857&0.6370& 0.2186
    & 0.0700& 0.0329& 0.9297& 0.8929&0.3386& 0.2484&0.4618\\
    &F1& 0.4685&0.8592&0.7684&0.3469&0.1283&0.0612&0.9401&0.9332&0.4907&0.3882&0.5385\\
    \midrule
    
    \multirow{2}{*}{STIL~\cite{gu2021spatiotemporal}}&
    R& 0.7143 & 0.8043 & 0.8848 & 0.6771 & 0.5457 & 0.5571 & 0.9393 & 0.8957 & 0.7200 & 0.5011 &0.7239\\
&F1&0.0922&0.5533&0.7240&0.4864&0.5199&0.4181&0.5921&0.7317&0.5099&0.4292&0.5057\\
    \midrule
    
    \multirow{2}{*}{MINTIME-CLIP-B~\cite{coccomini2024mintime}}
    &R&0.6964&0.9257&0.9116&0.1071&0.2293&0.1071&0.9968&0.8600&0.7371&0.5637 & 0.6135\\    &F1&0.5532&0.9297&0.9378&0.1827&0.3633&0.1827&0.9630&0.9087&0.8177&0.6988 & 0.6538\\
    \midrule
    
    \multirow{2}{*}{FTCN-CLIP-B~\cite{Zheng2021ExploringTC}}&
    R&0.8929&0.9786&0.6297&0.4043&0.1650&0.1557&0.9984&0.9971&0.6514&0.4060&0.6279\\    &F1&0.8130&0.9765&0.7664&0.5660&0.2802&0.2636&0.9858&0.9922&0.7775&0.5697&0.6991\\
    \midrule
    
    \multirow{2}{*}{TALL~\cite{xu2023tall}}&
    R&0.5714&0.7643&0.7957&0.6443&0.3486 &0.5629 &0.8514&0.9143 &0.6714&0.4438&0.6568  \\
    &F1&0.1730&0.7053&0.7957&0.6289&0.4498&0.5727&0.7403&0.86423&0.6469&0.5077& 0.6085\\
    \midrule 
    
    \multirow{2}{*}{XCLIP-B-FT~\cite{ni2022expanding}}&R& 0.8571 & 0.9543 & 0.7623 & 0.6586 & 0.3593 & 0.3700 & 0.9968 & 0.9900 & 0.7557 & 0.4978 & 0.7201 \\
&F1&0.7705&0.9632&0.8580&0.7814&0.5234&0.5291&0.9826&0.9879&0.8468&0.6548&0.7898\\
    \midrule
    
    \multirow{2}{*}{DeMamba-XCLIP-FT~\cite{chen2024demamba}}&R &0.9464 & 0.9843 & 0.9217 &  0.8243 & 0.5229 & 0.5400 & 0.9952 & 0.9914 & 0.7929 &  0.5788&0.8098 \\
&F1& 0.5000&0.9255&0.9240&0.8381&0.6559&0.6412&0.9228&0.9609&0.8192&0.6855&0.7873\\
    \midrule
    
    \multirow{2}{*}{ReStraV~\cite{interno2025ai}}&R&-&-&-&-&-&-&-&-&-&-&0.8200\\
    &F1&-&-&-&-&-&-&-&-&-&-&\underline{0.8980}\\
    \midrule
    
    \multirow{2}{*}{NSG-VD~\cite{zhang2025physics}}&R&0.9464&1.0000&0.9833&1.0000&0.9750&1.0000&1.0000&1.0000&0.9167&0.8917&\textbf{0.9713}\\
&F1&0.8983&0.8985&0.8839&0.8664&0.8897&0.8791&0.9057&0.8824&0.8397&0.8045&0.8745\\
    \midrule

    \rowcolor{mygray}&R&0.8750&0.9957&0.9942&0.9200&0.8850&0.8814&1.0000&0.9993&0.9929&0.9238&\underline{0.9467}\\
    % real 0.9985
     \rowcolor{mygray}
     \multirow{-2}{*}{\textbf{MPF(Ours)}}&F1&0.8099&0.9866&0.9913&0.9471&0.9333&0.9257&0.9874&0.9940&0.9851&0.8947&\textbf{0.9455}\\
    
    \bottomrule
    \end{tabular}
    }
    % \vspace{-10pt}
    \label{tab:genvideo_senie}
\end{table*}

\textbf{Dataset Specifications.} The training strategy strictly adheres to the `one-to-many' evaluation setting established by the GenVideo benchmark \cite{chen2024demamba}. Under this protocol, the training data is derived exclusively from real videos sourced from Youku-mPLUG and fake videos from the Pika or SEINE subset. Crucially, the input to the Micro-Temporal Fluctuation Branch is limited to pixel-wise signal features and contains no high-level semantic information.

% \textbf{Dataset Specifications.} \textit{1) For the Static Manifold Deviation Branch}: We leverage ImageNet as the primary representative of the on-manifold real-world data distribution ($\mathcal{M}_{real}$), while Stable Diffusion v1.4 from GenImage \cite{zhu2023genimage} is employed as the source of prototypical off-manifold synthetic samples. \textit{2)For the Micro-Temporal Fluctuation Branch}: the training strategy strictly adheres to the `one-to-many' evaluation setting established by the GenVideo benchmark \cite{chen2024demamba}. Under this protocol, the training data is derived exclusively from real videos sourced from YouTube and fake videos from the Pika subset. Crucially, the input to this branch is limited to pixel-wise signal features and contains no high-level semantic information.

\textbf{Training Configuration}. For the Static Manifold Deviation Branch, we only fine-tune the linear head. For the Micro-Temporal Fluctuation Branch, we apply LoRA to the terminal 1/4 layers of the MetaCLIP-v2 backbone. The specific hyperparameter settings are as follows: 1) Epochs: Both branches are trained for only 1 epoch. 2) Sequence Length ($L$): The maximum number of continuous frames for micro-temporal analysis is set to $L=8$. 3) LoRA Hyperparameters: We set the rank $lora_r=4$, the scaling factor $lora_\alpha=8$, and a dropout rate of 0.1.

\vspace{-1mm}
\subsection{Benchmark Evaluation on Legacy Datasets}
\vspace{-1mm}

To evaluate the effectiveness of our framework in filtering off-manifold samples common in early-stage generative videos, we first conduct assessments on the VidProm benchmark \cite{wang2024vidprom}. These legacy datasets are typically characterized by low frame rates (FPS), restricted resolutions, and significant compression artifacts,

The quantitative results are summarized in Tab.~\ref{tab:vidprom}. To ensure a fair and transparent comparison, all baseline performance metrics reported in the table are directly sourced from \cite{interno2025ai} and \cite{wang2024vidprom}. Our framework was trained using a balanced dataset of real and synthetic videos primarily sourced from Youku-mPLUG and Pika (GenVideo). As illustrated in the results, our framework achieves SOTA performance across nearly all VidProm subsets.

\vspace{-1mm}
\subsection{Evaluation on the GenVideo Benchmark}
\vspace{-1mm}

To further validate the generalization capability of our framework across the rapidly evolving spectrum of generative models, we conducted a `one-to-many' evaluation using the GenVideo benchmark \cite{chen2024demamba}. We maintain a transparent and rigorous comparison protocol, with all baseline metrics directly sourced from established evaluations in \cite{chen2024demamba} and \cite{zhang2025physics}. Due to our specific hierarchical framework, we explicitly omit the Average Precision (AP) metric from our comparative analysis. Apart from the NSG-VD method, which exclusively utilizes real-world training data from Kinetics-400, all other baselines and ours strictly adhere to the default training strategy prescribed by the GenVideo one-to-many protocol.

As detailed in Tab~\ref{tab:genvideo_pika}, with `PIKA' training condition, our framework achieves State-of-the-Art (SOTA) performance across both Recall and F1-score. And as shown in Tab~\ref{tab:genvideo_senie}, with `SENIE' training condition, our framework achieves the higher Recall while simultaneously maintaining a SOTA F1-score.

% To evaluate the efficacy of our framework against the next generation of synthetic media, we conducted a specialized assessment on the proposed High-Fidelity `In-the-Wild' (HF-ITW) dataset. This experiment is specifically designed to verify our performance on videos with high FPS and extreme fidelity.We benchmark our framework against several state-of-the-art VFM-based forensic models. For consistency, all compared baselines have undergone similar fine-tuning on the GenVideo Pika dataset. As demonstrated in Tab.~\ref{tab:HF-ITW}, our hierarchical framework achieves SOTA performance on the HF-ITW dataset, significantly outperforming single-branch VFM detectors.

\vspace{-1mm}
\section{Correlation Analysis: Impact of Temporal Fidelity on MPF Efficacy}
\vspace{-1mm}

The efficacy of MPF is intrinsically tied to both the temporal and spatial fidelity of the video. To quantify this relationship, we define a Composite Quality Score for each dataset subset, mapped from three key dimensions: FPS, Mbps, and spatial resolution.

As illustrated in Fig.~\ref{fig:outline_MPFefficacy}, we sorted all evaluated subsets by their composite quality score, with video quality gradually decreasing from left to right.

\textbf{High-Fidelity Scenarios (e.g., Sora, Pika, Moonvalley)}:
High-frequency sampling preserves the local smoothness of the latent trajectory $\Delta z$. This allows the Jacobian-driven homogeneity $\mathbf{J}_\mathcal{D}$ to manifest clearly without interference. In these scenarios, the MPF branch consistently achieves filtering accuracies exceeding 70\%.

\textbf{Low-Fidelity Scenarios (e.g., Lavie)}: As FPS and resolution drop, the increased temporal distance between frames violates the infinitesimal assumption of our Taylor expansion:$\Delta I \approx \mathbf{J}_\mathcal{D} \cdot \Delta z$. This leads to  `catastrophic perturbations', where subtle projection fluctuations are obscured by coarse motion artifacts and compression noise. Consequently, performance degrades significantly, occasionally falling toward random-guess levels.

\vspace{-1mm}
\section{Conclusion}
\vspace{-1mm}
In this paper, we introduced a hierarchical dual-path framework to address the forensic challenges posed by high-fidelity `on-manifold' synthetic videos. By defining MPF, we identified a universal computational fingerprint rooted in the deterministic reconstruction logic of generative decoders. Our framework employs a Static Manifold Sentinel to filter coarse spatial anomalies and a Micro-Temporal Microscope to expose structured homogeneity in high-fidelity samples.

\nocite{langley00}

\bibliography{example_paper}
\bibliographystyle{icml2026}

%%%%%%%%%%%%%%%%%%%%%%%%%%%%%%%%%%%%%%%%%%%%%%%%%%%%%%%%%%%%%%%%%%%%%%%%%%%%%%%
%%%%%%%%%%%%%%%%%%%%%%%%%%%%%%%%%%%%%%%%%%%%%%%%%%%%%%%%%%%%%%%%%%%%%%%%%%%%%%%
% APPENDIX
%%%%%%%%%%%%%%%%%%%%%%%%%%%%%%%%%%%%%%%%%%%%%%%%%%%%%%%%%%%%%%%%%%%%%%%%%%%%%%%
%%%%%%%%%%%%%%%%%%%%%%%%%%%%%%%%%%%%%%%%%%%%%%%%%%%%%%%%%%%%%%%%%%%%%%%%%%%%%%%
\newpage
\appendix
\onecolumn

\section*{Structure of Contents}
\begin{compactitem}
    \item \textbf{Section~\ref{sup:additional MPF Analysis}: Additional MPF Analysis:}
    \begin{compactitem}
        \item Section~\ref{sup:Evidence of MPF via Comprehensive Consistency Score}: Evidence of MPF via Comprehensive Consistency Score;
        \item Section~\ref{sup:QualitativeAnalysisMPF}: Qualitative Analysis of MPF Enhancement Candidates;
    \end{compactitem}
    \item \textbf{Section~\ref{sup:Branch Synergy Analysis}: Branch Synergy Analysis}
    \begin{compactitem}
        \item Section~\ref{sup:evalVFM}:Evaluation of VFM Backbones for the Static Sentinel;
        \item Section~\ref{sup:arch}: Architecture Ablation of the MPF Branch;
    \end{compactitem}
    \item \textbf{Section~\ref{sup:VideoQUalityAnalysis}: Video Quality Analysis and its Impact on MPF Performance}
    \begin{compactitem}
        \item Section~\ref{sup:QuantitativeQualityMetrics}: Quantitative Quality Metrics;
        \item Section~\ref{sup:VisualizingtheDetectionBoundary}: Visualizing the Detection Boundary.
    \end{compactitem}
\end{compactitem}

\section{Additional MPF Analysis}
\label{sup:additional MPF Analysis}

\subsection{Evidence of MPF via Comprehensive Consistency Score}
\label{sup:Evidence of MPF via Comprehensive Consistency Score}

The essence of Manifold Projection Fluctuations (MPF) resides in the micro-temporal discrepancies between consecutive frames. To mathematically capture these traces, we define a Comprehensive Consistency Score ($S_{total}$) that evaluates two fundamental dimensions of the frame residuals: Quantity Consistency ($C_{quant}$) and Spatial Distribution Consistency ($C_{spatial}$).

\begin{table*}[h!] % Use [h!] or similar for placement in appendix
  \centering
    \caption{Quantitative evaluation of MPF consistency metrics across real and synthetic datasets.}
    % \vspace{-5pt}
    \label{tab:consistencyscore} % Keep original label
  \footnotesize
  \setlength{\tabcolsep}{3pt}
  \begin{adjustbox}{width=\linewidth}
      \begin{tabular}{c|c|cccc|cccccccccc}
        \toprule
         &\multirow{2}{*}{Real}&\multicolumn{4}{c}{\textbf{VidProM}}&\multicolumn{10}{c}{\textbf{GenVideo}}\\
        \cmidrule(lr){3-6} \cmidrule(lr){7-16}
         
        & &Pika&VC2&T2VZ&MS &Sora&Morph Studio & Gen2 &HotShot&Lavie&Show\-1&Moon Valley&Crafter&Model Scope&Wild Scrape\\
        \midrule
        $C_{qty}$&0.3328&0.5261&0.9020&0.9838&0.9546&0.7916&0.9010&0.6102&0.7816&0.7402&0.7839&0.3613&0.8956&0.8907&0.5948\\
        $C_{spa}$&0.0788&0.1321&0.2656&0.3505&0.3344&0.2268&0.2556&0.1715&0.2509&0.2124&0.2264&0.0820&0.2610&0.2524&0.1731\\
        \bottomrule
      \end{tabular}
  \end{adjustbox}
    % \vspace{-10pt}
\end{table*}

\textbf{Consistency Metric Formulation.} To objectively evaluate different enhancement strategies, we define a Comprehensive Consistency Score ($\mathcal{S}_{cons}$). This metric measures the stability of forensic traces throughout a video segment. It is calculated as a weighted fusion of Quantity Consistency ($C_{qty}$) and Spatial Consistency ($C_{spa}$):$$\mathcal{S}_{cons} = w_1 \cdot C_{qty} + w_2 \cdot C_{spa}$$Quantity Consistency ($C_{qty}$): Computed using the inverse standard deviation of change ratios and mask densities across frames:
$$C_{qty} = \frac{1}{1.0 + \sigma(\text{change\_ratio}) + \sigma(\text{mask\_density})}$$Spatial Consistency ($C_{spa}$): Derived from the temporal stability of the change centroids $(x, y)$ and regional distribution (border vs. center):$$C_{spa} = \frac{1}{1.0 + \sigma(centroid_x) + \sigma(centroid_y) + \sigma(ratio_{border}) + \sigma(ratio_{center})}$$We set $w_1 = w_2 = 0.5$ in our studies to ensure that the chosen feature captures both the `how much' and the `where' of the micro-temporal signal.

As summarized in Tab.~\ref{tab:consistencyscore}, the quantitative evaluation of these metrics provides empirical proof for the theoretical divergence described in Sec.~\ref{sec:gunisnotenough}. The $C_{qty}$ score for real-world videos is significantly lower than that of any synthetic subset. This supports our observation in Sec.~\ref{sec:gunisnotenough}: in authentic videos, stationary backgrounds often achieve absolute pixel stability ($\Delta I \approx 0$), causing the change ratio to fluctuate wildly (high $\sigma$) when subtle physical noise or motion occurs.

Furthermore, the notably low $C_{spa}$ values for real videos confirm that authentic residual features are unstructured and heterogeneous, lacking a consistent spatial `anchor'. The Structured Homogeneity of MPF. In contrast, all synthetic subsets, even high-fidelity ones like Sora and Pika exhibit substantially higher consistency scores. This serves as a direct macroscopic manifestation of the Jacobian-driven homogeneity defined in Sec.~\ref{sec:gunisnotenough}. Unlike the chaotic fluctuations of physical sensors, the residuals of AI videos are forced into a structured and homogeneous pattern by the frozen decoder. This repetitive utilization of the same pixel perturbation patterns results in the high temporal and spatial stability captured by our microscope branch, providing a robust statistical boundary for forgery detection.

\subsection{Qualitative Analysis of MPF Enhancement Candidates}
\label{sup:QualitativeAnalysisMPF}

\begin{figure*}[h!]
    \centering
  \includegraphics[width = 0.9\textwidth]{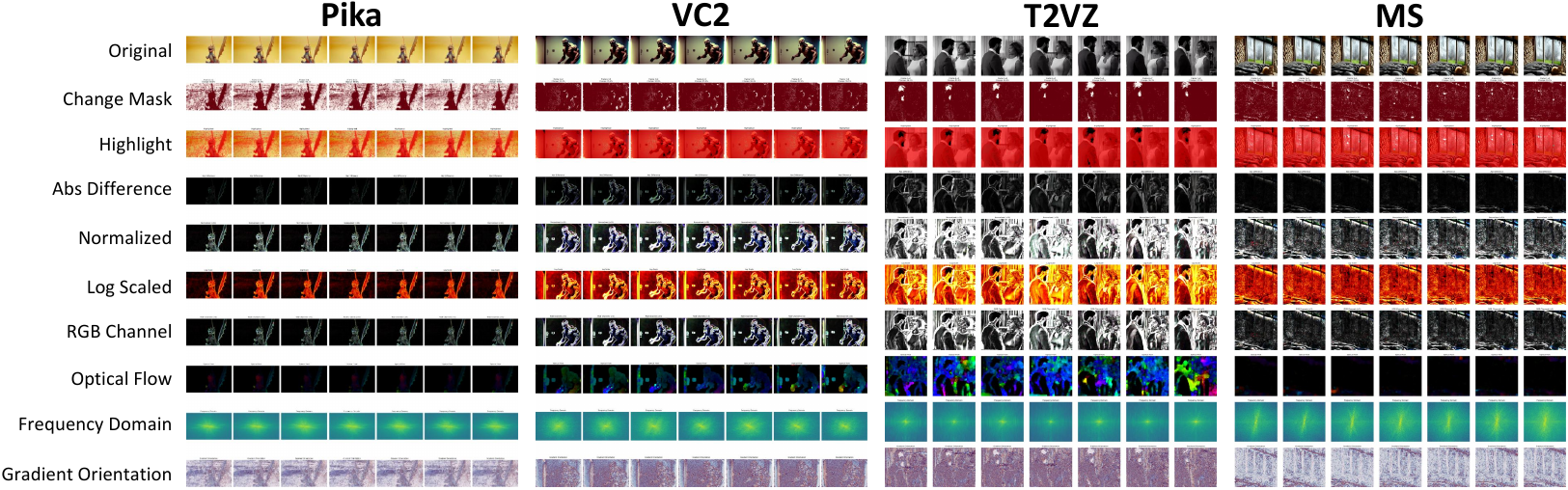}
  \caption{Visual manifestations of forensic features under different transformation strategies (VidProM).}
  \label{fig:MPFsamples}
  % \vspace{-10pt}
\end{figure*}

\begin{figure*}[h!]
    \centering
  \includegraphics[width = 0.9\textwidth]{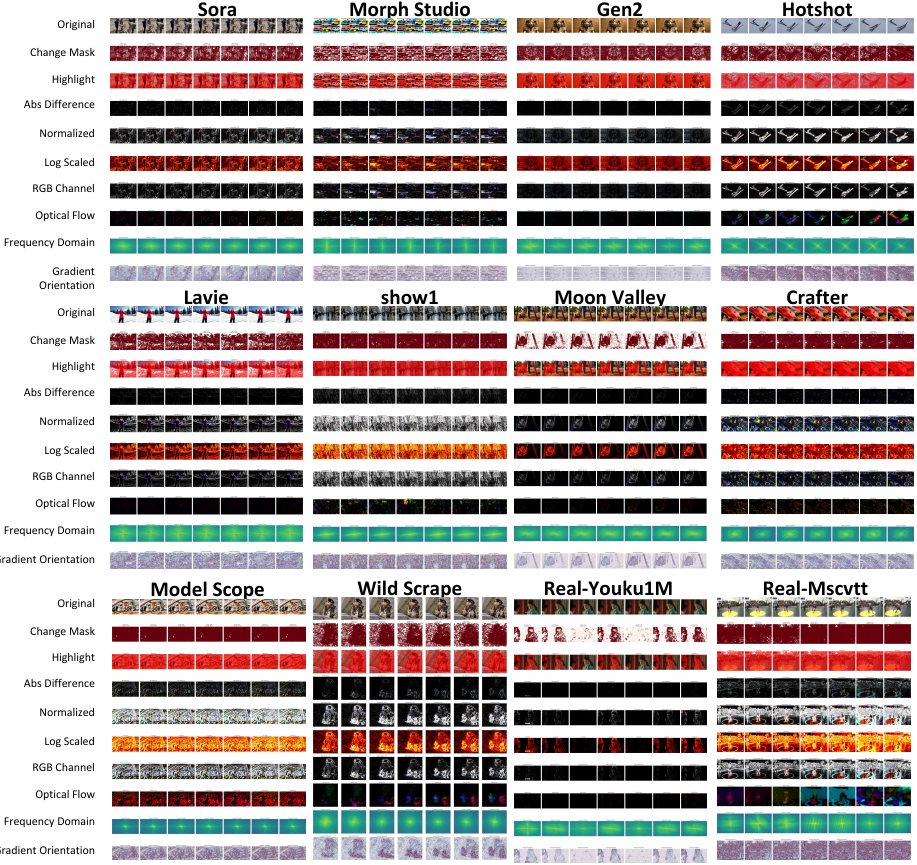}
  \caption{Visual manifestations of forensic features under different transformation strategies (GenVideo).}
  \label{fig:MPFsamples}
  % \vspace{-10pt}
\end{figure*}

We investigated nine distinct strategies to transform raw frame differences into forensic features. Our findings are categorized below based on their ability to preserve the MPF signal (ref to shown Fig~\ref{fig:MPFsamples}):

\begin{compactitem}
    \item \textbf{Binary-Centric Methods (Change Mask \& Highlighted)}: These methods focus purely on the quantity of changed pixels. While they provide high contrast for motion detection, they discard the internal texture required to identify manifold projection homogeneity. They fail to distinguish between the structured noise of an AI decoder and the random jitter of a physical sensor.
    \item \textbf{Transformation \& Motion Methods (Optical Flow, Frequency Domain, \& RGB Channels)}:
    \begin{compactitem}
        \item Optical Flow tends to over-smooth micro-fluctuations, effectively erasing the subtle Jacobian traces we aim to detect.
        \item Frequency Domain (FFT) is excellent at identifying structural periodicities but lacks a direct mapping to the magnitude of change, making it less effective for variable-quality videos.
        \item RGB Channel separation often dilutes the overall signal-to-noise ratio, making the MPF signal harder for the backbone to isolate.
    \end{compactitem}
    \item \textbf{Non-Linear Scaling Methods (Log Scale \& Gradient Orientation)}: Both methods show high potential in capturing both quantity and structure. Log Scale is particularly adept at balancing micro-changes and macro-motions, while Gradient Orientation excels at detecting unnatural texture shifts. However, they are computationally more intensive than linear normalization.
    \item \textbf{The Optimal Strategy - Clamped Normalization (Normalized $\times 10$)}: Defined as $\text{Input} = \text{Clamp}(|I_t - I_{t+1}| \times 10, 0, 255)$. This method emerged as the most effective lens for our `forensic microscope'.
\end{compactitem}

Through extensive sampling across all dataset subsets, we found that Normalized ($\times 10$) residuals offer the best performance for high-fidelity (on-manifold) detection for three reasons: \textbf{1)} Scaling the difference by a factor of 10 elevates the micro-fluctuations into a range where the pre-trained VFM weights can perceive the low-entropy homogeneity characteristic of AI manifold projections. \textbf{2)} Unlike masks, this feature preserves the relative intensity of pixel changes. This allows the model to differentiate between the high-entropy stochastic noise of real sensors and the deterministic fluctuations of AI decoders. \textbf{3)} The clamping operation ensures that extreme outliers (e.g., sudden flashes or fast motion) do not overwhelm the micro-temporal signal, maintaining a stable input distribution for the LoRA-tuned layers.

As shown in Tab~\ref{tab:mpfvsothermpf}, we conducted an extensive ablation study to evaluate five distinct transformation strategies for MPF signal enhancement. Each variant was trained on a balanced set of real videos from Youku-mPLUG and high-fidelity synthetic samples from Pika. To provide a comprehensive measure of detection sensitivity, we calculated the Mean Recall across all evaluated subsets.

The results indicate that the Normalized MPF strategy achieves SOTA performance, significantly outperforming other heuristic or raw-pixel approaches. This superiority stems from its ability to mitigate content-dependent intensity variations (e.g., varying lighting or complex textures) while effectively amplifying the underlying Jacobian-driven structural patterns. By standardizing the residual distribution, the Normalized strategy allows the Microscope branch to focus exclusively on the low-entropy homogeneity of the manifold projection, thereby maximizing its discriminative power against `on-manifold' forgeries that otherwise blend into the natural data distribution.

\begin{table*}[h!] % Use [h!] or similar for placement in appendix
  \centering
    \caption{Ablation study of MPF enhancement strategies.}
    % \vspace{-5pt}
    \label{tab:mpfvsothermpf} % Keep original label
  \footnotesize
  \setlength{\tabcolsep}{3pt}
  \begin{adjustbox}{width=\linewidth}
      \begin{tabular}{c|cccc|cccccccccc|c}
        \toprule
         \multirow{2}{*}{Aggregate Architecture} &\multicolumn{4}{c}{\textbf{VidProM}}&\multicolumn{10}{c}{\textbf{GenVideo}}&\multirow{2}{*}{Avg.}\\
         \cmidrule{2-5} \cmidrule(lr){6-15} 
         
         &Pika&VC2&T2VZ&MS &Sora&Morph Studio & Gen2 &HotShot&Lavie&Show\-1&Moon Valley&Crafter&Model Scope&Wild Scrape&\\
        \midrule
        Frequency Domain&0.9998&0.9996&0.9913&0.8115&0.8750&0.9986&0.9978&0.8200&0.8814&0.7771&1.0000&1.0000&0.9943&0.8370&0.9274\\
        Change Mask&0.9999&0.9996&0.9928&0.8123&0.8036&0.9857&0.9920&0.8157&0.8986&0.7900&1.0000&1.0000&0.9857&0.8535&0.9235\\
        Log Scale&0.9999&0.9998&0.9960&0.8115&0.8014&0.9857&0.9986&0.8600&0.8986&0.8043&1.0000&0.9993&0.9886&0.8337&0.9270\\
        Optical Flow&0.9998&0.9997&0.9945&0.8117&0.7845&0.9846&0.9988&0.8062&0.8914&0.7963&1.0000&1.0000&0.9941&0.8321&0.9190\\
        \midrule
        \rowcolor{mygray}
        Normalized (Ours)& 1.0000 & 0.9997 & 0.9938 & 0.8115&0.8393&0.9900&0.9978&0.8571&0.9079&0.8429&1.0000&0.9993&0.9943&0.8348&0.9334\\
        \bottomrule
      \end{tabular}
  \end{adjustbox}
    % \vspace{-10pt}
\end{table*}

\section{Branch Synergy Analysis}
\label{sup:Branch Synergy Analysis}

In this section, we provide an in-depth analysis of the individual contributions of our dual-path architecture. We first evaluate the `Sentinel' capability of different Vision Foundation Models (VFMs) in the first stage, followed by an architectural ablation of the "Microscope" branch to justify our design choices.

\subsection{Evaluation of VFM Backbones for the Static Sentinel}
\label{sup:evalVFM}

The effectiveness of the Static Manifold Deviation Branch is intrinsically tied to the VFM's ability to define a rigorous boundary for the real-world manifold $\mathcal{M}_{real}$. We compared four prominent VFMs across different release eras to observe how the evolution of pre-training scale affects forensic ·pruning‘ performance:

Evaluated Models \& Release Timeline:
\begin{compactitem}
    \item \textbf{CLIP (Feb 2021)}: The foundational vision-language baseline.
    \item \textbf{SigLIP (Mar 2023)}: Optimized via sigmoid loss for better localized feature alignment.
    \item \textbf{SigLIP-2 (Feb 2025)}: Refined multi-modal distillation for sharper perceptual boundaries.
    \item \textbf{PE (Apr 2025)}: provides specific alignment tuning methods to make these features accessible for tasks ranging from zero-shot classification to dense spatial prediction and multimodal language understanding.
    \item  \textbf{MetaCLIP-v2 (Jul 2025)}: Our primary backbone, offering the most dense and detailed manifold projection.
\end{compactitem}

\begin{table*}[h!] % Use [h!] or similar for placement in appendix
  \centering
    \caption{Interception efficacy of the Static Manifold Deviation branch across VFM generations..}
    % \vspace{-5pt}
    \label{tab:VFMcom} % Keep original label
  \footnotesize
  \setlength{\tabcolsep}{3pt}
  \begin{adjustbox}{width=\linewidth}
      \begin{tabular}{c|cccc|ccccccccccc}
        \toprule
         &\multicolumn{4}{c}{\textbf{VidProM}}&\multicolumn{10}{c}{\textbf{GenVideo}}
         &\multirow{2}{*}{Real}\\
         \cmidrule(lr){2-5} \cmidrule(lr){6-15} 
         
        \multirow{-2}{*}{Method}&Pika&VC2&T2VZ&MS &Sora&Morph Studio & Gen2 &HotShot&Lavie&Show\-1&Moon Valley&Crafter&Model Scope&Wild Scrape\\
        \midrule
        CLIP&0.9637&0.7764&0.8535&0.1093&0.5857&0.9200&0.8942&0.3471&0.6293&0.3100&0.9840&0.9886&0.8557&0.6090&0.9955\\
        SigLIP&0.3867&0.9057&0.9987&0.8278&0.5000&0.5771&0.3949&0.8429&0.7921&0.7729&0.3626&0.6767&0.2571&0.6982&0.9659\\
        SigLIP2&0.7540&0.9896&0.9967&0.9928&0.3214&0.9314&0.9036&0.6914&0.7971&0.8600&0.9968&0.9850&0.8129&0.7258&0.9929\\
        PE&0.9857&0.9992&0.9926&0.8115&0.4286&0.9657&0.9884&0.8114&0.8936&0.7900&1.0000&0.9964&0.9543&0.8216&0.9984\\
        MetaCLIP-v2&0.9784&0.9982&0.9971&0.8762&0.6871&0.9914&0.9884&0.8300&0.9036&0.8714&1.0000&1.0000&0.9543&0.7148&0.9799\\
        \bottomrule
      \end{tabular}
  \end{adjustbox}
    % \vspace{-10pt}
\end{table*}

As shown in Tab~\ref{tab:VFMcom}, we observe a direct correlation between the VFM’s release date (and consequent training scale) and its interception rate. Early models like the original CLIP and SigLIP, constrained by limited data volume, tended to collapse complex real-world structures into lower-dimensional feature representations. This  `structural neglect' allowed many synthetic samples to masquerade as natural data within the model's diffuse decision space. In contrast, MetaCLIP-v2 leverages internet-scale pre-training to internalize a vastly more comprehensive and dense representation of $\mathcal{M}_{real}$.

This massive exposure effectively `smooths out' the gaps in the model's understanding of natural textures and physical laws, enabling it to achieve the highest "early pruning" rate on benchmarks like VidProm—intercepting over 97\% of samples in Stage I. These results suggest that as foundation models scale, they become inherently more sensitive to `disturbances' in the natural distribution, allowing the Sentinel to function as a highly efficient first-pass filter for off-manifold content.

\subsection{Architecture Ablation of the MPF Branch}
\label{sup:arch}

A critical design question for the Micro-Temporal Fluctuation Branch was how to aggregate temporal information across the $L=8$ frames. We compared our `Simple Concatenation' strategy against more complex temporal modeling modules.

\begin{compactitem}
    \item \textbf{Temporal Transformer}: Appending a multi-head self-attention layer across the $L$ [CLS] tokens to model long-range dependencies.
    \item \textbf{LSTM}: Employing recurrent units to capture sequential frame-to-frame transitions.
    \item \textbf{Simple Concatenation (Ours)}: Directly flattening and concatenating the $L$ [CLS] tokens into a single forensic vector.
\end{compactitem}

\begin{table*}[h!] % Use [h!] or similar for placement in appendix
  \centering
    \caption{Ablation study of temporal aggregation strategies within the MPF branch.}
    % \vspace{-5pt}
    \label{tab:AblTemporalAgg} % Keep original label
  \footnotesize
  \setlength{\tabcolsep}{3pt}
  \begin{adjustbox}{width=\linewidth}
      \begin{tabular}{c|cccc|cccccccccc|c}
        \toprule
         \multirow{2}{*}{Aggregate Architecture} &\multicolumn{4}{c}{\textbf{VidProM}}&\multicolumn{10}{c}{\textbf{GenVideo}}&\multirow{2}{*}{Avg.}\\
         \cmidrule{2-5} \cmidrule(lr){6-15} 
         
         &Pika&VC2&T2VZ&MS &Sora&Morph Studio & Gen2 &HotShot&Lavie&Show\-1&Moon Valley&Crafter&Model Scope&Wild Scrape\\
        \midrule
        Temporal Transformer&0.9994&0.994&0.9930&0.8116&0.6964&0.9714&0.9964&0.8143&0.8836&0.7829&1.0000&0.9979&0.9771&0.8249&0.9102\\
        LSTM&1.0000&0.9997&0.9924&0.8117&0.7679&0.9786&0.9978&0.8286&0.8979&0.7971&1.0000&0.9979&0.9871&0.8271&0.9203\\
        \midrule
        \rowcolor{mygray}
        Simple Concatenation (Ours) & 1.0000 & 0.9997 & 0.9938 & 0.8115&0.8393&0.9900&0.9978&0.8571&0.9079&0.8429&1.0000&0.9993&0.9943&0.8348&\textbf{0.9334}\\
        \bottomrule
      \end{tabular}
  \end{adjustbox}
    % \vspace{-10pt}
\end{table*}

The `Simplicity is Stability Finding: Consistent with our claims in Sec~\ref{sec:microtemporalbranch}, our empirical analysis reveals that simple concatenation provides superior generalization performance, particularly for OOD (Out-of-Distribution) datasets. Complex modules like Temporal Transformers tend to develop an inductive bias toward specific macro-motion patterns (e.g., a specific way a character walks in the training set).

The `Simplicity is Stability' Finding. Consistent with our theoretical framework in Sec~\ref{sec:microtemporalbranch}, our empirical analysis reveals that simple concatenation provides superior generalization performance, particularly for Out-of-Distribution (OOD) datasets.

As shown in Tab.~\ref{tab:AblTemporalAgg}, complex modules like Temporal Transformers tend to develop an inductive bias toward specific macro-motion patterns (e.g., the specific gait of a walking character). This causes the model to ·overfit’ to the semantic content of the training set rather than the underlying physics. In contrast, by using simple concatenation, we prevent the network from learning high-level temporal relationships that might lead to semantic leakage. This ensures the branch remains focused on the Jacobian-driven homogeneity of the micro-residuals—the `computational DNA' that remains invariant regardless of the specific action occurring in the video.

\section{Video Quality Analysis and its Impact on MPF Performance}
\label{sup:VideoQUalityAnalysis}

The efficacy of Manifold Projection Fluctuations (MPF) is intrinsically tied to the precision of the video's spatio-temporal signal. To rigorously evaluate this relationship, we investigate how the "perceptual fidelity" of different generative models affects the detectability of their latent projection traces.

\subsection{Quantitative Quality Metrics}
\label{sup:QuantitativeQualityMetrics}

We introduce a Composite Quality Profile to quantify the generation characteristics of each model. This profile is derived from three fundamental technical dimensions:

\begin{compactitem}
    \item \textbf{Temporal Frequency (FPS)}: Determines the density of sampling along the latent trajectory $\Delta z$. Higher FPS ensures the validity of the first-order Taylor expansion used in our MPF derivation.
    \item \textbf{Information Density (Bitrate/Mbps)}: Reflects the degree of compression. Low bitrates introduce stochastic quantization noise that can obscure the structured homogeneity of the Jacobian $\mathbf{J}_\mathcal{D}$.
    \item \textbf{Spatial Resolution (N)}: Defines the observation space. A higher $N$ provides more pixel-wise constraints to resolve the $M$-dimensional manifold projection.
\end{compactitem}

\begin{table*}[h!] % Use [h!] or similar for placement in appendix
  \centering
    \caption{Technical profiles and Composite Quality Scores of generative subsets.}
    % \vspace{-5pt}
    \label{tab:qualityScore} % Keep original label
  \footnotesize
  \setlength{\tabcolsep}{3pt}
  \begin{adjustbox}{width=\linewidth}
      \begin{tabular}{c|cccc|cccccccccc}
        \toprule
         &\multicolumn{4}{c}{\textbf{VidProM}}&\multicolumn{10}{c}{\textbf{GenVideo}}\\
        \cmidrule(lr){2-5} \cmidrule(lr){6-15}
         
        &Pika&VC2&T2VZ&MS &Sora&Morph Studio & Gen2 &HotShot&Lavie&Show\-1&Moon Valley&Crafter&Model Scope&Wild Scrape\\
        \midrule
        $C_{fps}$&20&10&4&8&30&8&24&8&15.52& 8& 50& 9.81& 8& 14.41\\
        $C_{bit}$&1.7& 3.1& 1.5&0.8&10.86& 4.13& 3.59& 1.51& 2.37& 0.38& 3.5& 12.43& 1.58& 3.16\\
        $C_{N}$&684769& 163840&262144&65536& 
            1430178& 589824& 808151& 258048& 163840& 184320& 795648&623575&691630& 382382\\
        \bottomrule
      \end{tabular}
  \end{adjustbox}
    % \vspace{-10pt}
\end{table*}

As summarized in Tab.~\ref{tab:qualityScore}, these scores vary significantly across the landscape of generative models, from high-fidelity engines like Sora and Pika to earlier architectures like Lavie.

\subsection{Visualizing the Detection Boundary}
\label{sup:VisualizingtheDetectionBoundary}

To visualize the operational boundaries of our framework, we provide a Quality-Accuracy Correlation Map (Fig. S8). In this chart:

\begin{compactitem}
    \item X-axis: Represents the Composite Quality Score, indicating the overall technical fidelity of the video.
    \item Y-axis: Represents the Stage 2 Detection Accuracy, measuring the success rate of the MPF branch.
    \item Bubble Size: Corresponds to the Remaining Samples (the volume of videos that successfully passed the Stage 1 Sentinel). This represents the "hard-to-detect" on-manifold samples that specifically require the Microscope's intervention.
\end{compactitem}

\begin{figure*}[h!]
    \centering
  \includegraphics[width = 0.7\textwidth]{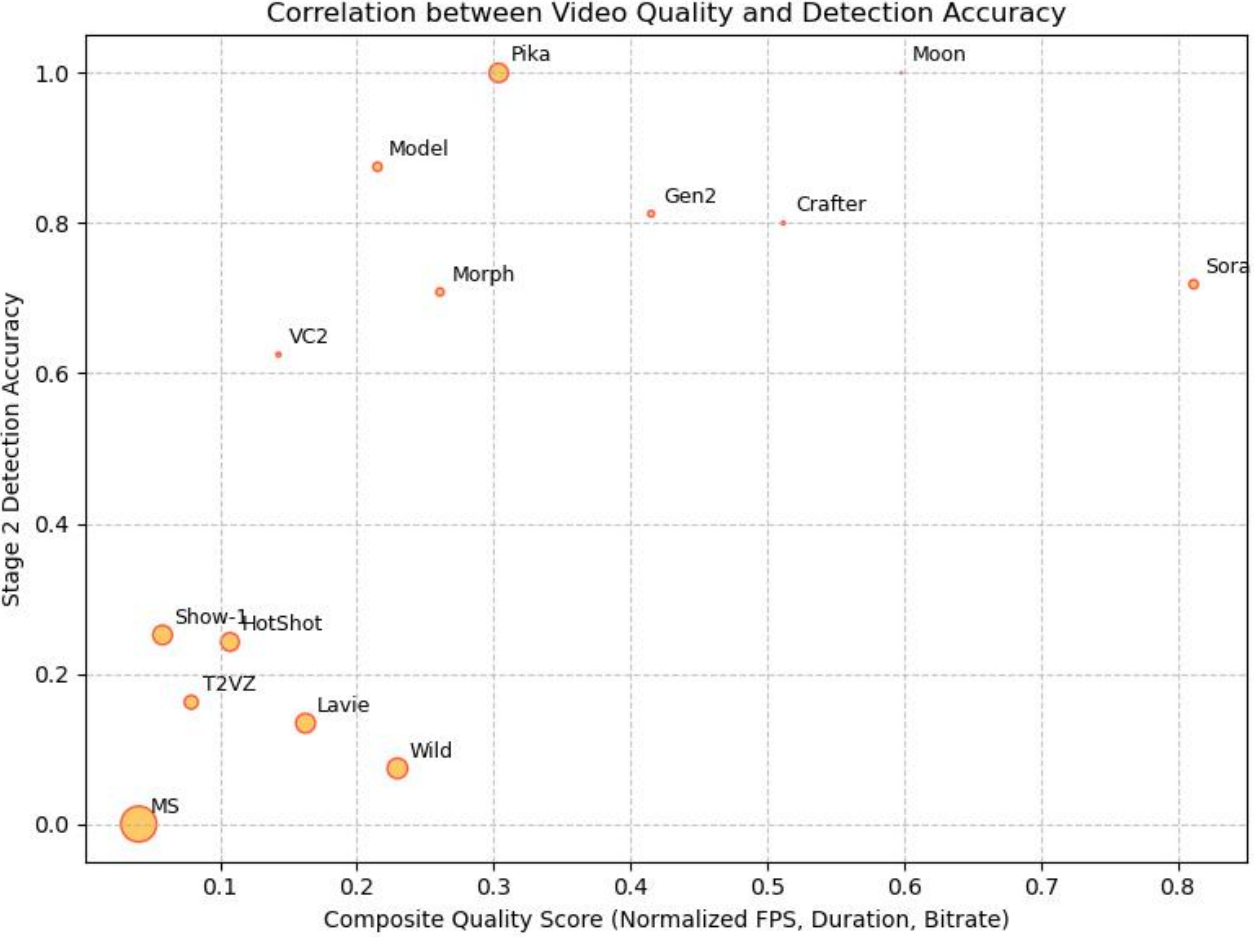}
  \caption{Correlation between Video Quality and Stage 2 Detection Accuracy. The orange bubbles represent different generative subsets, with size indicating the volume of `on-manifold‘ samples that bypass Stage 1.}
  \label{fig:qualityVsaccuracy}
  % \vspace{-10pt}
\end{figure*}

\textbf{Finding 1: The `Fidelity Dividend.} A strong positive correlation is observed between $Q_{comp}$ and detection accuracy. As shown in Fig. S8, high-fidelity models (located at the right of the plot) yield the most reliable MPF signatures. This confirms our theoretical derivation: when the video quality is high, the Jacobian-driven homogeneity ($\mathbf{J}_\mathcal{D}$) is less obscured by random artifacts, allowing for near-perfect forensic identification.

\textbf{Fingding 2: The Sentinel-Microscope Handover.} The bubble sizes reveal that high-quality subsets often have larger `remaining sample' volumes. This indicates that Stage 1 (the VFM-based Sentinel) struggles to prune these high-fidelity samples because they reside deep within the natural data manifold. However, their high Y-axis position proves that Stage 2 (the Microscope) effectively `resolves' these cases, precisely where Stage 1 is most vulnerable.

\textbf{Finding 3: The Operational Floor.} For subsets with extremely low $Q_{comp}$ (the left side of the plot), the accuracy exhibits a downward trend. This is attributed to the violation of the infinitesimal assumption in our Taylor expansion; coarse motion and low frame rates introduce `catastrophic perturbations' that degrade the micro-temporal signal.

\end{document}